\documentclass{article}
\usepackage[final, nonatbib]{nips_2016}
\usepackage[utf8]{inputenc}
\usepackage{amsmath}
\usepackage{graphicx}
\usepackage{amssymb}
\usepackage[T1]{fontenc}    
\usepackage{hyperref}       
\usepackage{url}            
\usepackage{booktabs}       
\usepackage{amsfonts}       
\usepackage{nicefrac}       
\usepackage{microtype}      
\usepackage[english]{babel}

\usepackage[
backend=bibtex,
style=numeric,
sorting=ynt
]{biblatex}
\title{Recent Advances in Neural Program Synthesis}

\author{
    Neel Kant \\
    Machine Learning at Berkeley  \\
    UC Berkeley  \\
    \texttt{kantneel@berkeley.edu}  
}

\begin{document}

\maketitle
\setlength{\parindent}{5ex}
\nocite{*}

\begin{abstract}
    In recent years, deep learning has made tremendous progress in a number of fields that were previously out of reach for artificial intelligence. The successes in these problems has led researchers to consider the possibilities for intelligent systems to tackle a problem that humans have only recently themselves considered: program synthesis. This challenge is unlike others such as object recognition and speech translation, since its abstract nature and demand for rigor make it difficult even for human minds to attempt. While it is still far from being solved or even competitive with most existing methods, neural program synthesis is a rapidly growing discipline which holds great promise if completely realized. In this paper, we start with exploring the problem statement and challenges of program synthesis. Then, we examine the fascinating evolution of program induction models, along with how they have succeeded, failed and been reimagined since. Finally, we conclude with a contrastive look at program synthesis and future research recommendations for the field. 
\end{abstract}

\section{Introduction}
As an overall field, program synthesis can be defined as the task of developing an algorithm that meets a specification or a set of constraints. The constraints are what serve to define the algorithm, since they impose criteria for correctness. These conditions may include runtime properties such as speed and space complexity and practically always include examples of correct input and output. For computer science students, this idea is quite familiar, as exams almost definitely include questions in which you are either asked to straightforwardly write pseudocode \textit{e.g. Write a ternary quicksort algorithm} or fill in blanks of a partially filled algorithm. 

There are a great deal of applications for program synthesis. Successful systems could one day automate a job that is currently very secure for humans: computer programming. Imagine a world in which debugging, refactoring, translating and synthesizing code from sketches can all be done without human effort. Furthermore, consider that there are other problems that computer programs do not directly solve, but could be framed as programming questions. Theorem proving in mathematics and physics is an example of a  task that require humans to produce new insights based on previously existing principles. A complete program synthesis system could run a program \textsc{ProveOrDisprove} on some predicate and perform what most would call a very creative task. While computer vision aims to automate a sophisticated perceptual system of living creatures, program synthesis is a field that aims to solve reasoning, logic and automation itself. 

These incredibly powerful applications of program synthesis make the field an enticing one to study, yet, it will likely take several decades before results of that magnitude can be realized. Likewise, deep learning has gathered a great deal of interest lately, and is being tried as a tool at pretty much every cognitive task. This paper serves to highlight the recent progress made in the intersection of these two fields. As we will see, there are achievements and challenges that ought to both, humble us and inspire us to continue our pursuits.

\begin{figure}[h!]
\centerline{\includegraphics[width=\linewidth]{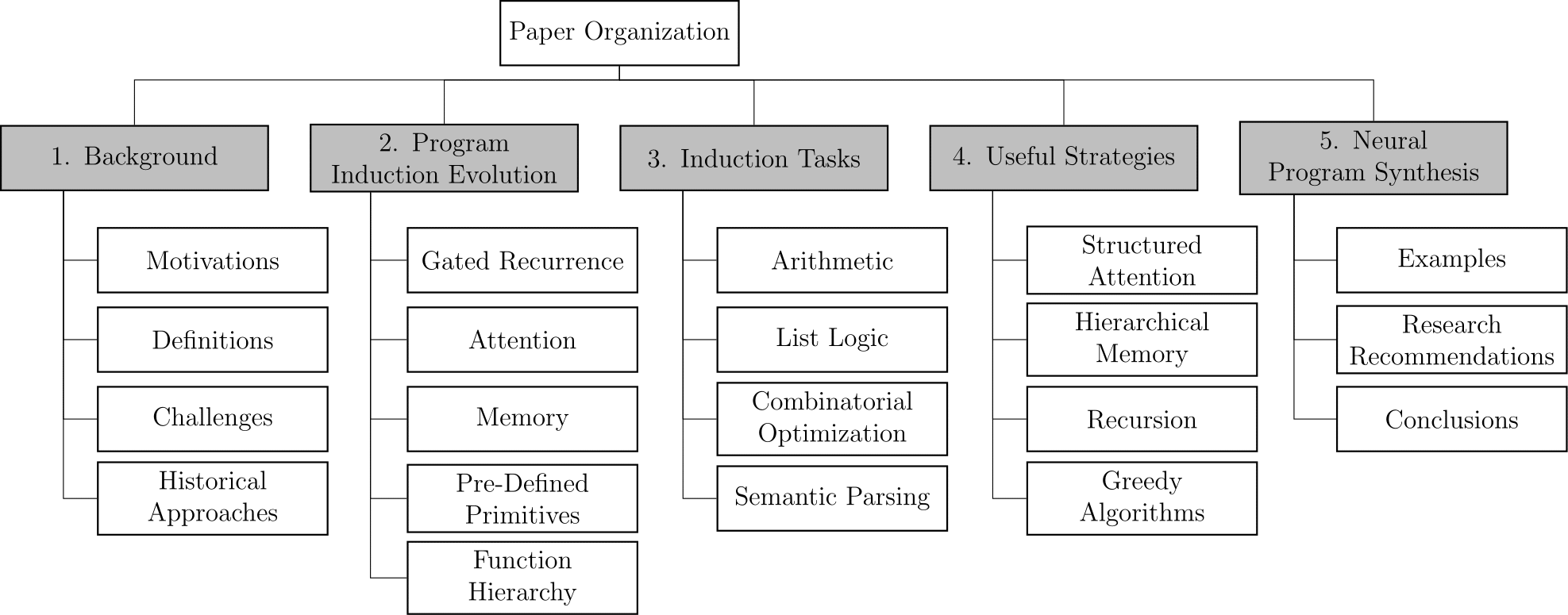}}
\caption{\textit{A hierarchical overview of this paper's organization.}}
\end{figure}

\subsection{Black Box Programming}
Suppose I had a black box, which if you provided enough constraints for a program, it would synthesize a program for you. Examples of constraints are (input, output) pairs which the program must satisfy or a natural language description of the program. The correct program $\mathcal{P}$ takes input $x$ and produces $y = \mathcal{P}(x)$. Consider two different kinds of black box program synthesizers.
\begin{enumerate}
    \item The black box produces $\hat{P}$ and allows you to see its mechanics. Then, when test inputs are given, the black box returns $\hat{y} = \hat{P}(x)$. This is called \textbf{program synthesis} because the model has explicitly returned a program.
    \item The black box produces $\hat{P}$ but you cannot tell what the mechanics of the program are. Then, when test inputs are given, it will perform operations invisible to you and eventually return an answer $\hat{y}$. This is called \textbf{program induction}, since the model learns to \textit{mimic} the program rather than explicitly return it. 
\end{enumerate}
$\hat{P}(x)$ from black box $(1)$ is clearly easier to recognize as \textit{always, sometimes or never} correct. Traditional program synthesis has primarily dealt with this paradigm since it has a high degree of interpretability. However, as we will discuss, this approach hit many walls in earlier decades. Black box $(2)$ should remind us of modern deep learning models. Convolutional Neural Networks (CNNs) are hard to interpret and they do not present clear-cut logic for how to classify images. But while CNNs are powerful in their ability to "perceive" and "intuit" the correct answers, they also fall prey to tricky "edge cases" known as adversarial examples \cite{adexamples}. We should be wary of any program induction black box since establishing guarantees on correctness is made far more difficult. 

We should ask ourselves, what sort of black box would the human program synthesizer be? As an example, we have two people, Alice and Bob. Alice presents Bob a series of (input, output) pairs: $(3, 9), (5, 25), (9, 81) \dots$. Once Bob is ready, Alice asks what is $P(10)?$ Bob gives the correct answer: 100. Alice could then also ask what is $P(24)$ and Bob may give the correct answer again. At what point can Alice truly be sure that Bob has actually synthesized $P$ and not some close approximation $\hat{P}?$ It starts to become a philosophical question. Generalization for program induction models, such as Bob can be practically guaranteed by trying out program sizes of much larger size e.g. $P(6.022 * 10^{23})$ or things we could consider tricky for other reasons, e.g. $ P(-6), P(\pi) \dots$. 

Of course, Bob could also simply say that $P(x) = x^2$ and what that entails for all $x \in \mathbb{R}$. Then Bob would have performed \textit{program synthesis} and Alice would not have to ask any more test questions. This example should show how program synthesis is generally preferable to induction, but also appears intuitively far harder for deep learning models to learn. 

\subsection{A Unique Challenge for Deep Learning}
The task of program synthesis is quite different from others that deep learning has excelled at. Some of the breakthroughs of modern deep learning include object and speech recognition, language translation and Atari game playing. What's interesting about each of these problems and generally the things that deep learning succeeds at is that the input space is continuous. Furthermore, data is distributed across these input spaces, and only when we as humans look at an aggregate of data can we distinguish \textit{entities with meaning}. For example, in computer vision, an image is vectorized in $\mathbb{R}^d$ when composed of $d$ pixels. That vector has a continuous domain and what we would discern as a cat or pedestrian has its presence distributed across numerous pixels. Only when those pixels and their data is \textit{aggregated and abstracted} is object recognition possible. The same is true of audio applications, since anything we would discern as phonologically meaningful takes multiple timesteps of audio waveforms to capture. Arguably, language is a discrete domain space since one cannot interpolate between words. But deep learning can teach models semantics by learning continuous domain embeddings of words, and in that space, semantics is distributed across all dimensions of the embedding vector.  

Deep learning is intuitively compatible for these problems because the knowledge of deep learning models is itself distributed throughout. Convolutional filters, recurrent hidden states, and nonlinearities on operations parameterized by millions of weights are well-suited for transforming \textbf{continuous, high-dimensional, and distributed} representations of discrete entities. In literature, this distributed approach to knowledge is known as \textbf{connectionist AI} \cite{syncon}. Its primary advantage is that having the output of the model be a differentiable function of its parameters means that gradient based optimization is possible. This optimization against distributed processing units is what enables (hopefully) straightforward learning regimes where averaged loss and error both decline stably and dependably.

Program synthesis is unlike the aforementioned problems. This is because the atomic structures of programs have inherent meaning in themselves, and thus there is no continuous domain space. While this may seem just like natural language processing, it is quite different because the range of vocabulary in programs is far smaller and also holds abstract value that does not embed well into a manifold. Not only this, but generating programs also usually entails arithmetic and logical operators, and operators take arguments. Training data usually consists of ordered pairs of (input, output) with the task being to induce the program's functionality into the model itself. The downside of connectionist AI in this setting is that distributing knowledge of programming constructs is often ineffective, and deducing the knowledge of a model after it has trained is practically impossible. Since a concrete program can rarely, if ever, be extracted from a connectionist deep learning model, generalization is impossible to guarantee and unsurprisingly, the models fail to execute the program on inputs of arbitrary size.

\subsection{The Current State of the Art}
But program synthesis is a problem studied by artificial intelligence researchers for decades. The original approach completely avoided continuous representations because during the founding of the discipline in the 1960's, the computational world was far more constrained. Instead, early program synthesis researchers focused on \textbf{symbolic AI} \cite{syncon}. The fundamental building block for this paradigm was the notion of a symbol, which when combined, form expressions. To be meaningful these expressions follow a \textbf{context free grammar} and form a \textbf{(domain specific) language.} Since there is no pre-defined limitation on what symbols can represent and expressions can be constructed and manipulated in arbitrarily complicated ways, symbolism is fully expressive and theoretically unrestricted. By simply guiding the construction of expressions by rules and heuristics, i.e. a program, expressions can be constructed which resemble programs themselves. 

The downside of symbolic AI approaches are that learning is not a straightforward task. Since expressions, and hence programs, are constructed by \textit{purely logical} operations rather than by \textit{differentiable functions}, gradient based optimization is not available. In fact, symbolic AI learning is often framed as a constraint satsifaction problem. The standard approach is to use rule-based methods to construct a program and to search as narrow a slice of the complete program space as possible. For reference, one of the most common tools used today are \textbf{satisfiability modulo theories} (SMT) solvers. The search, and hence, learning algorithm concludes when all of the constraints are met, which are typically just that the assembled program works correctly on the given (input, output) pairs. While the the learning process may seem limited, symbolic AI systems have an undeniable advantage in terms of generalization and provable correctness. Furthermore, modern computer hardware is advanced enough that SMT solvers are very fast and can solve nontrivial problems very efficiently.

\begin{figure}[h!]
\centerline{\includegraphics[width=0.8\linewidth]{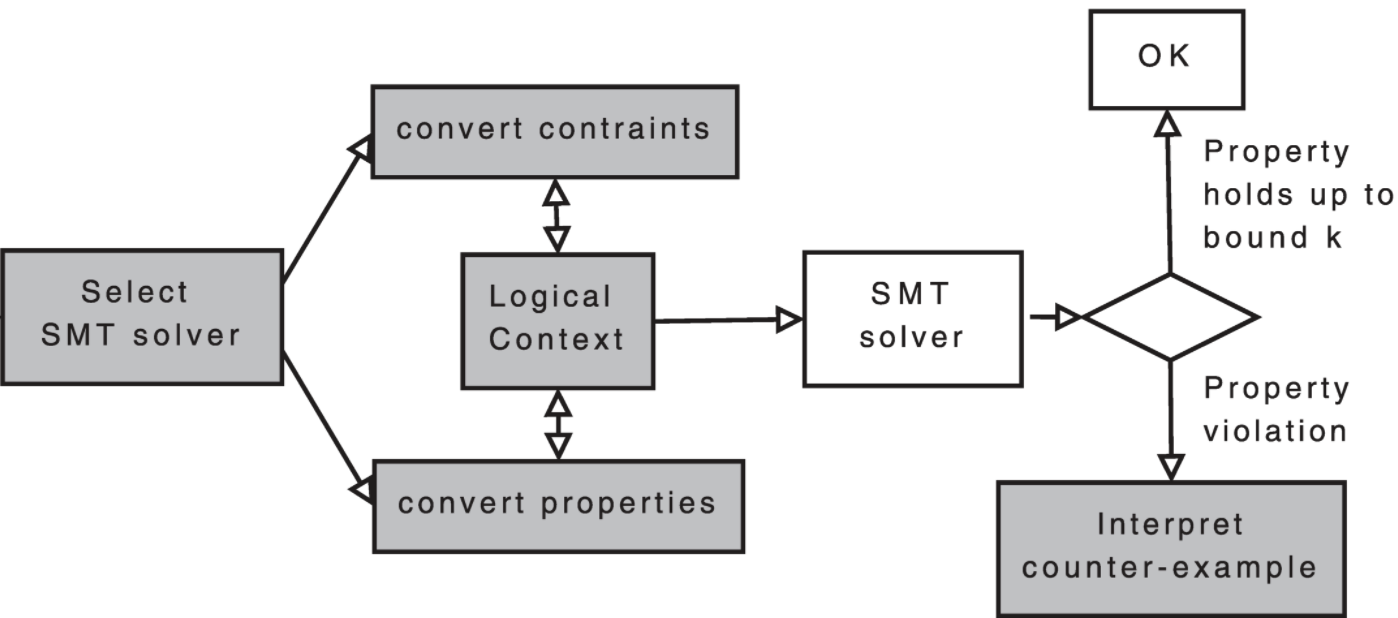}}
\caption{\textit{For each SMT solver, the program specification is interpreted into constraints and properties in some \textbf{theory} \textit{e.g. first-order logic, bitvector arithmetic}. Then a satisfiability search and verification procedure is used to return a valid program or a declaration of unsatisfiability.}}
\end{figure}

Though there are still some firm purists in both, the symbolic and connectionist AI camps, the consensus is that neither approach is likely to solve program synthesis in isolation. What is important is that the two paradigms have \textbf{complementary} strengths and weaknesses in their abilities to \textit{represent} and \textit{learn} knowledge. Thus, it is natural to believe that a \textbf{hybrid system} is the most promising path forward. We note that while humans are unique in their command over symbols, our information processing machinery reduces to a connectionist model of neural matter. The more successful neural programming models mimic this setup, where symbols are built up from embeddings of the program state space. 

\section{Neural Program Induction Models}
A variety of model paradigms and architectures have been proposed to tackle the wide-ranging challenges in neural program induction. While the models tend to share certain base properties which allow them to be categorized, they are quite creative in their details. This specificity can be used to explain why certain models may succeed at a task while others will fail. These details are also what give clues as to what sorts of properties are generally desirable and/or effective in tackling neural program induction tasks. 

\subsection{Intro - Recurrent Models}
Recurrent neural networks (RNN) are distinctive for their intuitive fit for sequential data. This makes them the only realistic deep learning choice for a variety of tasks including time-series analysis, audio processing, and various natural language processing tasks. A major practical breakthrough for RNNs was the creation of gated recurrence which is nowadays most often implemented with the Long Short-Term Memory (LSTM) cell.  

Given data embeddings e.g. \cite{glove}, LSTM networks are effective at classification and regression tasks such as image captioning. These networks often exploit an encoder-decoder framework, in which an LSTM network is first tasked with iteratively encoding data as it is passed in. The encoded data is then passed to another LSTM network (or the original network is starts a decoding phase) whereby the data is decoded and outputs are produced.  


RNNs are also a natural fit for programming tasks since both, inputs and outputs, are of variable size in program induction. If the task is program synthesis, then the output size cannot even be inferred from the input, and so a natural method of computation is to produce the output one token at a time, updating the internal state in the process. Some tasks can also be built on natural language inputs, and so having a network of RNNs working together is a natural idea. As we will see, some models are augmented with external memory resources, and so the RNN can issue requests in different time steps. 

The following sections show some of the \textbf{key evolutionary steps} that program induction models have taken. This progression did not happen chronologically, but I present them in this order because of how I perceive their \textbf{effective capacity.} The criteria for this includes measures of generalization ability, adaptability to different tasks, abstraction usage, and interpretability.  

\subsection{Convolutional Recurrence}
Recall that convolutional neural networks (CNN) are a core class of deep learning models. A convolution is a mathematical operation in which two functions are 'blended' together across a domain. In deep learning, the two functions are 1), the value of the input data and 2) a filter which scans over the data. CNNs are powerful tools to process inputs that exist in tensors of order greater than 1 because the convolution operation picks up information in spatial neighborhoods of the data.

One particularly successful neural program induction model is the Neural GPU \cite{ngpu}, which is recurrent, but each "timestep" involves a gated convolutional operation. Input data is first embedded into a 3D tensor called the "internal state" of the model which is the analog of a hidden state of an RNN. At timestep $t$, the model acts by applying gated convolutional operations by means of a Convolutional Gated Recurrent Unit (CGRU). The mechanism is identical to a GRU, except the matrix multiplications on vectors are replaced with convolutional actions on a 3D tensor. 

Importantly, there is an "update" and "reset" gate which can be used to preserve long-term information much like an LSTM. The output of a CGRU also always has the same shape as the input. At timestep $t+1$, the Neural GPU applies the same CGRU operations on CGRU($t$), and in this way is both convolutional and recurrent. The point is that for some input size $n$, the Neural GPU simply applies its CGRU operation $n$ times. After these operations, the internal state of the model is decoded to produce the output for the program.  If the model succeeds for small $n$, then the hope is that it will also work for larger problem sizes, simply by repeatedly applying the operations.

Convolutional recurrence is a clever idea, but it has a shortcoming in the Neural GPU implementation. At timestep $t$, all of the information on the problem from timesteps $1 \dots t-1$ is stored in the internal state and there is no way to view the prior internal states separately. This puts a lot of faith on the model's internal state to contain all of the necessary information to proceed in the next timestep. 

\subsection{Attention and Pointer Networks}
The issue discussed in the prior section can be addressed with a mechanism known as attention. Attention is valuable because it lets a decoder access the information of each encoder state without producing a bottleneck. Neural machine translation techniques as in \cite{join-align} use this to consider each input token \textit{individually} when formulating the output. More broadly, however, attention distributions can be thought of as simply generating a \textit{non-negative distribution that sums to 1}, e.g. standard probability distributions. In the encoder-decoder paradigm, this means that the decoder can selectively attend to each of the encoder states to the degree that is most helpful. In program induction and synthesis, attention can also be used to relax discrete operations, such as selection.

\begin{figure}[h!]
\centerline{\includegraphics[width=0.8\linewidth]{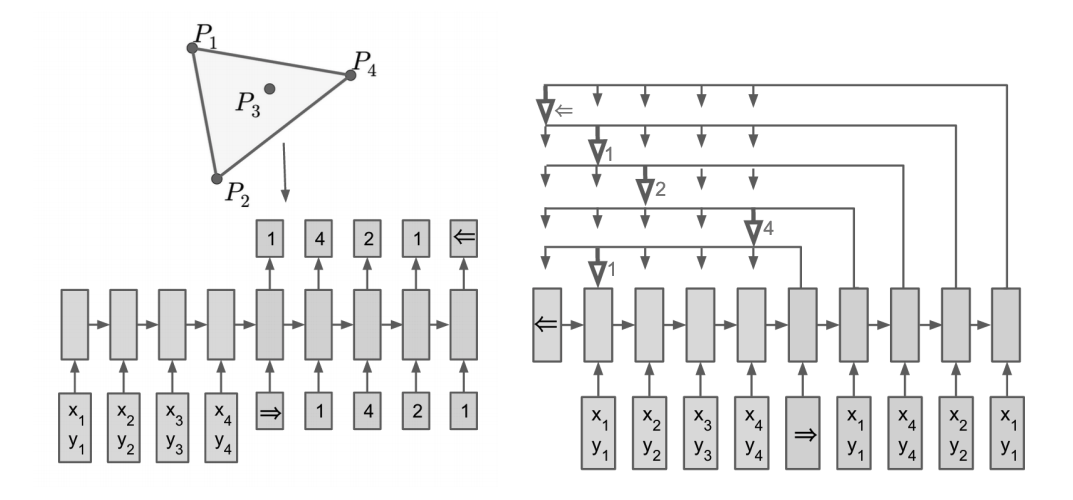}}
\caption{\textit{A visual representation of pointers referring to objects from the input dictionary.}}
\end{figure}

Pointer Networks (Ptr-Nets) \cite{ptr} are perhaps the most straightforward nontrivial application of attention-equipped RNNs to neural program synthesis. In the paper, the authors used three examples of problems that involve points in a coordinate space: Finding a convex hull, Delaunay triangulation and the Traveling Salesman Problem. The operations of the Ptr-Net are summarized as follows:

\begin{enumerate}
    \item Sequentially feed inputs into the model, which is behaving as an encoder. The encoder hidden states $e_1 \dots e_k$ are all recorded.
    \item Once input is completed, the model sequentially generates attention distributions on $\{e_1 \dots e_k, d_t\}$ with $d_t$ being the current decoder hidden state. These distributions serve as soft-selection during training and each one can be used in loss functions and gradient descent.
\end{enumerate} 

Importantly, the Pointer network is able to solutions on problems with \textit{greater input size than those that it was trained on.} This is a straightforward sign of generalization, but it is easily inferrable that with input sizes orders of magnitude greater, the performance would deteriorate considerably. This reminds us of the crux of the problem with neural program induction models. Though they induce a program in the size regime within which they are trained, it is \textit{nearly impossible to guarantee performance} on greater problem sizes and \textit{verify corner cases.}

\subsection{Attention with Memory}
The Neural Turing Machine (NTM) \cite{ntm} introduced the notion of augmenting a neural network with external memory. In effect, the neural network now acts as a controller, issuing read and write commands to the memory, rather than having to use its own parameters as the main memory. The external memory takes the shape of a matrix $M \in \mathbb{R}^{m \times n}$

It is not immediately intuitive how a vector output from an RNN controller can be interpreted as a read or write command. Once again, attention mechanisms prove useful. The attention is applied with two addressing schemes in mind: 
\begin{enumerate}
    \item \textbf{Content Based} Addressing: For a read vector $v \in \mathbb{R}^n$, return a result $u \in \mathbb{R}^m$ where $u$ represents the similarity of the contents each memory slot and the read vector
    \item \textbf{Location Based} Addressing: The read/write vector $v \in \mathbb{R}^m$ represents the attention distribution over memory indices which to read/write. 
\end{enumerate}

\begin{figure}[h!]
\caption{\textit{Attentive steps to get the read output $\mathbf{w}_t$ via content addressing. Controller (LSTM) outputs are fully differentiable.}}
\centerline{\includegraphics[width=0.7\linewidth]{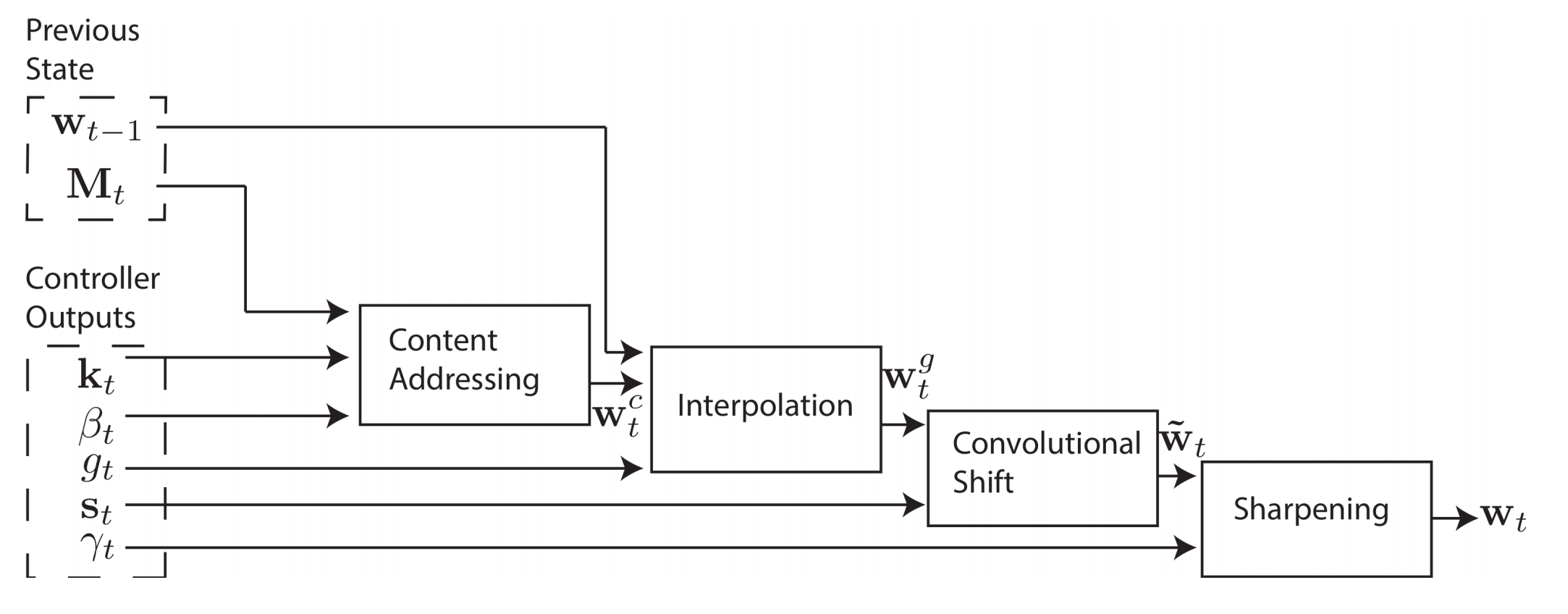}}
\end{figure}

These two mechanisms are simple and elegant. In each timestep, the NTM takes input, produces read and write commands, which, when taken with the RNN controller output interact with memory. However, the results that are demonstrated in the paper are noticeably elementary compared to Pointer Networks, especially with regards to problem size. The tasks themselves are not algorithmically complex either. These results are thought provoking because despite considerably more flexibility and expressiveness, the NTM is \textbf{significantly harder to train.} 

Graves et. al. took the ideas of the Neural Turing Machine a few steps further with the Differentiable Neural Computer (DNC) \cite{dnc}. The DNC can be trained with multiple read/write heads and also has additional data regarding its memory. In particular, these are matrices that detail: 
\begin{enumerate}
    \item \textbf{Temporal Linkages}: Information about the relative order with which the memory was written to. This should allow the controller to infer relationships between data since temporal connections are common in algorithms.  
    \item \textbf{Memory Usage}: Information about whether the index in memory is holding useful information. This should theoretically inform and simplify controller choices for reading and writing. 
\end{enumerate}

\begin{figure}[h!]
\caption{\textit{Depiction of the DNC as described in \cite{dnc} Note the multiple read vectors in \textbf{b} and the linkages described in \textbf{d}}}
\centerline{\includegraphics[width=0.85\linewidth]{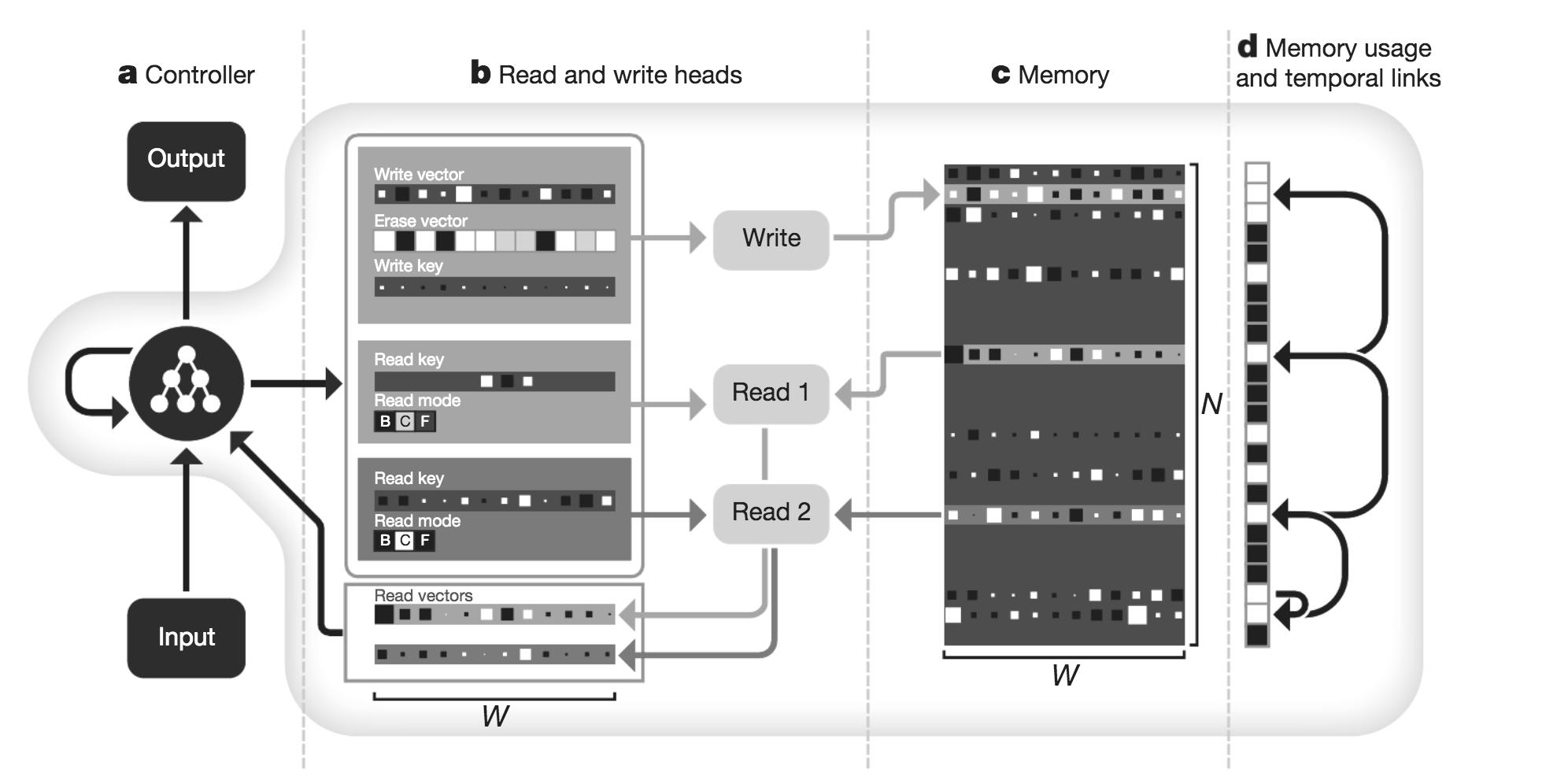}}
\end{figure}

The DNC still utilizes an RNN controller as its main core and uses attention-based addressing techniques. In the paper, the researchers demonstrated greater functionality with regards to task learning than the NTM. However, like the NTM, the DNC appears to be substantially harder to effectively train than simpler models like Pointer Nets. 

Extending models with external memory was a groundbreaking idea because of increased \textbf{expressibility} of induction models. In fact, the NTM and DNC are both \textbf{Turing Complete}, meaning that if given enough time steps and sufficient memory size, they can perform any set of instructions that a standard computer can. However expressive as they are, the models are far from possessing the requisite degree of \textbf{learnability} for such complex tasks. It appears that without tethering the models to any discretized, non-alterable operations, supervised gradient-based learning becomes prohibitively hard to utilize.

\subsection{Memory and Pre-Defined Primitives}
All the previous models have been \textit{purely connectionist} in their design. Now, we will explore two models, Neural Programmer \cite{nprog} and Neural RAM \cite{nram} that can only apply \textbf{explicitly defined transformations} on data.  In particular, the controllers do not issue direct read/write instructions to memory, and instead, perform one of several possible unary/binary operations on data. These are basic arithmetic (e.g. add, multiply), logic (e.g. equal comparison, less than) and aggregation (min, max) operators. The models are designed to run for more timesteps than are necessary to merely describe the problem. This gives the models a chance to \textbf{compose} these basic operations and create complex programs. 

In particular, the archetypal components are: 
\begin{enumerate}
    \item \textbf{RNN controller} that takes sequential inputs from (a) outside of the controller and/or (b) The memory unit - automatically delivered in an embedded form
    \item Learned functions that generate \textbf{attention distributions} over (a) the pre-defined operations to perform and (b) the data on which to perform those operations.
    \item A \textbf{memory unit} from which data is read written to. This can also serve as the designated output location (Neural RAM). 
\end{enumerate}

There are some interesting differences between the two models. The Neural Programmer is actually designed to take \textbf{natural language inputs.} Of course, these are embedded by an RNN controller and so become vectorized, but the model still needs to learn the semantics of English. Likewise, Neural Programmer has modules that let it perform database-type operations on its memory, and it can return multiple elements from the database. In this sense, Neural Programmer is designed to be an \textit{automatic Question-Answering system} that learns the latent programs required to answer its questions. Hence, the solutions may be compositional, but require fewer steps than there are tokens in the questions.

\begin{figure}[h!]
\caption{\textit{Schematic for the Neural Programmer \cite{nprog}. The input is a sequence of tokens and so takes multiple time steps to run.}}
\centerline{\includegraphics[width=0.7\linewidth]{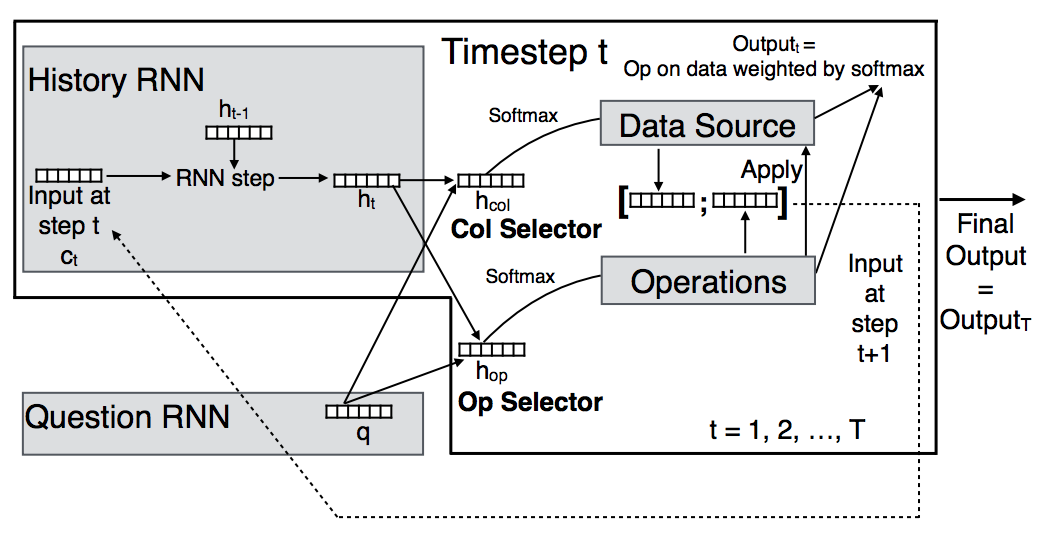}}
\end{figure}

On the other hand, Neural RAM tries to specialize in creating \textbf{highly compositional} programs. Since the model's 14 pre-defined modules are very atomic, the programs can take hundreds of timesteps to complete. For example, permuting a list takes $\mathcal{O}(n^2)$ operations when elements are only allowed to be moved one index at a time. So moving something $n$ indices to the left is exactly a composition of $n$ calls to "move left." Finding the maximum of $n$ numbers is also a composition of $\mathcal{O}(n)$ calls to "max($a$, $b$)." 

The good news is that compared to NTM and DNC, Neural RAM can create coherent programs that span a much \textbf{higher number of timesteps}. This is most likely because the operations are not "fuzzy," even if attention is applied to select them in varying proportions. When supervised with error backpropagation, the model will theoretically then know that one of the primitive operations was the correct one, rather than relying only on what the content of memory should have been written with (as is the case with NTM, DNC). The model also gets to choose \textbf{when to terminate} the program with a sigmoidal \textit{end} unit that, when exceeding a threshold, stops the model entirely. The caveat to the good news is that the tasks themselves are not that much more complicated compared to the NTM, and Neural RAM doesn't even attempt the graph problems that DNC has shown some skill at. 

\subsection{Function Hierarchy}
While Neural Programmer and Neural RAM control symbolic modules, the operations are quite elementary. Likewise, the training data for all of the aforementioned models is just (input, output) pairs. As mentioned in Section 1.2, these are \textit{not setups that favor deep learning models.} In particular, the supervision signal is weak and training data is prone to be overfitted against. 

The Neural Programmer-Interpreter (NPI) \cite{npi} seeks to address these issues by increasing \textit{model complexity} and \textit{supervision granularity.} An important advancement in the model is the flexibility to let functions call sub-functions in \textbf{new stack frames.} This is implemented by resetting the hidden state of the RNN controller to zeros in the new frame, and giving the embedded program, arguments and environment as the input. 

These stack frames can be terminated and control is returned to the caller frame by using a sigmoidal \textit{end} just like in Neural RAM. When a sub-function ends, the hidden state of the controller prior to calling the sub-function is restored. When the top-level function ends, then the entire model stops executing. 

The model components still resemble the hybrid models but have some important distinctions:
\begin{enumerate}
    \item\textbf{ RNN controller} that takes sequential state encodings built from (a) the world environment (changes with actions), (b) the program call (actions) and (c) the arguments for the called program. The entirety of the input is fed in the first timestep, so every action by the NPI creates an output that is delivered as input.
    \item Functions that select
    \begin{itemize}
        \item An \textbf{attention distribution} over which of the pre-defined programs to run and with what arguments. The programs are stored in a dictionary where the attention vector is matched against a key, and the program embedding is stored as the value for the key. 
        \item A sigmoidal "return to caller" unit to terminate the current stack frame
    \end{itemize}
    \item A memory block which forms a scratchpad. It is accessed through pointers which only exist at discrete locations. The NPI can move the pointers or write at the pointers' location as part of its most basic function \textsc{ACT}.
\end{enumerate}

\begin{figure}[h!]
\caption{\textit{NPI \cite{npi} scratchpad (left) and stack execution traces (right). Note: the lowest level of instructions are, in fact, \textsc{ACT} but made interpretable for the reader.}}
\centerline{\includegraphics[width=0.8\linewidth]{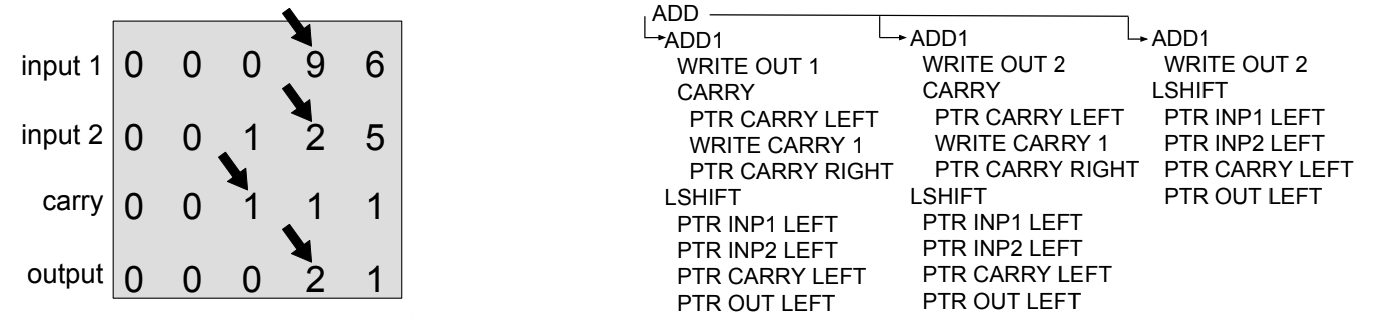}}
\end{figure}

The key idea is that the NPI is capable of \textbf{abstraction} and \textbf{higher-order} controls over the program. When a new function is called with arguments, it is expressed in an embedding vector. The embedding vector is constructed by combining embeddings of arguments, the world environment and context that would inform future calls to sub-functions. This is very different from Neural RAM, where the modules are selected by using attention, but the controller hidden state is the \textit{only place where information can be embedded}.  
Furthermore, the NPI's supervision is also enhanced because the researchers train the model with full execution stack traces. That means that every single function call in every stack frame is checked against a reference program, and this builds a richer supervision signal with which the NPI can train. This allows for solid training of the NPI architecture, but it is also a significant hurdle if the end goal is program induction for tasks that \textit{we do not know how to do efficiently.} Whereas training with input-output pairs can potentially allow cultivation of induction models that find short-cuts in program space that discrete language could not express, full execution trace supervision prevents this. The main desire that remains of the NPI is then that it can effectively learn with much weaker supervision strength. 

One approach that presents a solution comes from Neural Program Lattices (NPL) \cite{npl}. The model is an extension of the NPI which explicitly creates a stack frame hierarchy, and is also designed to work with \textbf{less supervision}. In particular, the architecture is designed with the challenge that often times, training data only has low-level instructions noted, and the abstractions are not explicitly defined or known. It would be preferable if models could learn to utilize a stack for function calls without training data that shows how to use the stack. The data would retain all of the instructions, except where new frames are pushed and popped from the stack, and a new neural network module uses dynamic programming estimate the likelihood of which stack frame the program exists in. The important result is that NPL is able to achieve similar performance to the NPI with this reduced supervision strength, and this represents another incremental step towards practical neural program induction. 

\subsection{Summary}
In this section we explored a number of evolutionary steps that neural program induction models have taken and give one or more models that epitomize each step in the evolution. A recurring theme is a balancing act between \textbf{expressiveness} and \textbf{trainability}. Increasingly complex models are \textit{theoretically} capable of accomplishing more, but unless requisite steps are taken to keep training effective, this promise remains unmaterialized. Many models (e.g. Neural GPU, Neural RAM, NPI) also require many random restarts to find a suitable parameter initialization. 

\section{Analysis of Model Performance on Induction Tasks}
This section serves to document the progress made on four examples of tasks which are arranged in increasing overall difficulty. The difficulty stems from the level of control flow required, the runtime complexity of the target program (if a man-made one even exists), and the availability of training data. 

\subsection{Arithmetic}
The Neural GPU (2.2) won great praise because of its ability to perform binary addition and binary multiplication quite well. In particular, the model can be trained on a curriculum of challenges up to around 20 digits in length, and that is sufficient to generalize to problems of thousands of digits in length. One caveat though is that the model cannot perform nearly as well when the same sum is represented in decimal notation. Furthermore, back from Section 1.1, we must ask ourselves, can we be sure that the Neural GPU does not suffer from some corner cases that trick it into messing up thousand-digit addition?

In \cite{engpu}, the researchers performed a more in-depth look at the performance of the Neural GPU. They found that the model cannot generalize in a complete manner because it is possible to manufacture highly structured inputs that fool it. For example, the model can correctly solve the multiplication problem $2 \times 2$ which has size $n=1$. However, they show that the model fails the same problem represented as $00\dots002 \times 00\dots002$ which has size $n=1000$. Another example of a failure mode is addition that requires a large number of consecutive "carry" operations. This is not very surprising, since there is no built-in notion of a carry operation in the model architecture. One cannot simply point to a collection of parameters and claim that this does the carry operation. It seems that the Neural GPU is more of a mental mathematician than a principled problem solver. 

\begin{figure}[h!]
\centerline{\includegraphics[width=0.8\linewidth]{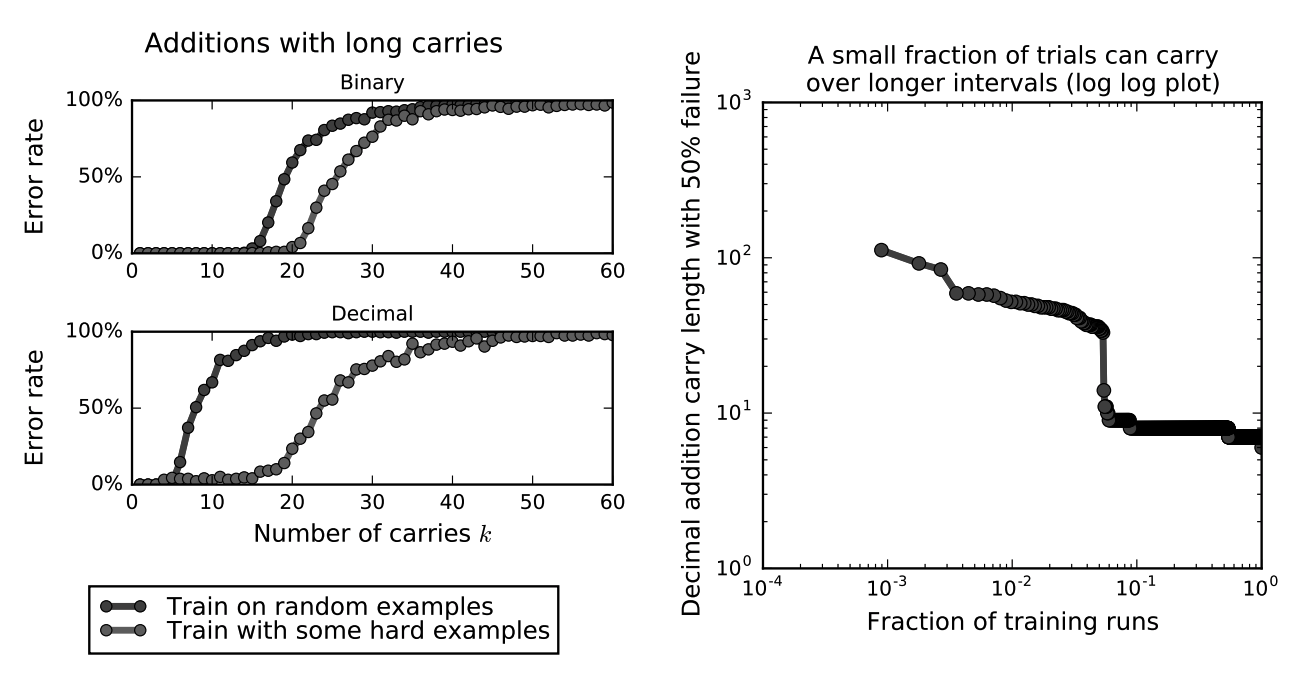}}
\caption{\textit{Despite specialized curriculum learning, the Neural GPU cannot generalize the notion of "carrying." (Left). A separate issue is that different parameter initializations make a large difference in overall performance (Right).} }
\end{figure}

We can compare the Neural GPU to the NPI and NPL (2.6). These models have also been tested on the task of addition, and can do so with decimal digits perfectly fine and in the same problem size regime. Because these models also implement a function hierarchy, they can generalize in a more trustworthy fashion. Namely, one of their abstract functions can be \textsc{Add} which may call \textsc{AddOne} which may call \textsc{Carry} if necessary. While it is true that some call of single-digit addition may glitch and not end up calling the required carry operation, there is no reason to suspect that the Neural GPU holds an advantage in this matter. Furthermore, we shall see in Section 4.3 that models that have function hierarchies can be made to implement recursion, and this can provide \textit{provable guarantees on generalization.} 

\subsection{List Logic}
The NTM (2.4) was the first new-wave neural program induction model to try tasks of this nature. Its achievements of copying and sorting lists of lengths around 20 were very notable in the program synthesis community (The paper currently has 487 citations). The paper draws comparisons between performance of the NTM and a regular LSTM network on these tasks, and rightfully shows that the addition of external memory greatly improves performance for LSTM networks. 

Neural RAM (2.5) is able to do copying, merging, permuting, reversing on list lengths of up to around 50. As mentioned in the description for the model, this is impressive because the induced program is composed of hundreds of timesteps and still manages to perform the task correctly. However, Neural RAM was not tested on list sorting, which would have been a good point of comparison against NTM.

On the point of list sorting, the NPI once again casts a shadow on these two aforementioned models. The paper specifically tests the model on an implementation of Bubblesort, and this serves to showcase the power of functional hierarchy. Sorting is a task that is easily specified in terms of input-output pairs, but there exist exponentially different implementations of the sorting procedure on one list, most of which are not systematic at all. The fuzzy attention based logic of the NTM does not give much confidence in terms of what sort of directed control flow the model is using, and whether it can efficiently do so for larger list lengths. NPI, on the other hand, uses a nicely structured set of commands, where \textsc{BubbleSort} calls \textsc{Bubble} which can call \textsc{CompSwap} \dots which will call \textsc{Act}. The supervision method of full execution traces also lets the NPI learn a specific sorting algorithm rather than an "intuitive" procedure. 

\subsection{Combinatorial Optimization} 
Combinatorial Optimization is a category of problem which requires \textit{optimizing a function} over a combination of \textbf{discrete objects} and the solutions are constrained. Examples include finding shortest paths in a graph, maximizing value in the Knapsack problem and finding boolean settings that satisfy a set of constraints. Many of these problems are \textbf{NP-Hard}, which means that no polynomial time solution can be developed for them. Instead, we can only produce approximations in polynomial time that are guaranteed to be some (hopefully constant) factor worse than the true optimal solution. They work by using heuristics which take an exponential action space and reduce it considerably. 

This field is ripe for neural program induction to tackle. It is conceivable that through distributed, embedded representations of the problem, better heuristics can be found that defy simple description. This idea was first tested with Pointer Networks, whose attention mechanism proved useful for the problem setup. Namely, the outputs are necessarily selected from a hand-picked dictionary and in combinatorial optimization, this dictionary is entirely specified by the input. Using the Traveling Salesman Problem as an example, the input dictionary is a list of vertex coordinates, and the output constructs a path through each of those vertices. Pointer networks succeed at problem sizes of moderate size, creating reasonable approximations, but they fail at higher sizes in a critical way. Not only are they unable to create competitive solutions, the outputs do not constitute valid solutions (e.g. a cycle that repeats some vertices and skips others). 

Likewise, the DNC (2.4) can utilize its memory structures to try and solutions to combinatorial optimization problems. The paper demonstrates the ability to find shortest paths in graphs, but the problem sizes are not very large. It can be inferred that the data structures of the DNC are not fully utilized to grasp the full structure of the problem, since training a fully differentiable model with external memory is quite difficult. In both the cases of the Pointer Network and DNC, supervised training with cross entropy loss was used. 

\subsection{Semantic Query Parsing}

In this problem setup, the biggest difference comes from the format in which the data is presented to the model. It is still (input, output) pairs, but now, the input is in the form of a query in a language, either human or programming in nature. This problem can blend the line between program synthesis and induction, because if the program to induce is a translating one (e.g. English to Python), then induction results in program synthesis. As mentioned earlier, recurrent models are essentially a prerequisite for handling tasks of query parsing. 

One model that was developed specifically for semantic query parsing and execution is the Neural Programmer (2.5). Another interesting example involves Pointer Networks \cite{seqsql}, where the objective is to synthesize SQL queries from natural language questions. In this scenario, there is the additional challenge of collecting a high-quality dataset for the task. Finding publicly available relational tables is not easy, and finding questions and SQL queries that are applicable for the tables is even harder. Oftentimes, the only realistic solution is to crowd-source this data online from sources like Amazon Mechanical Turk and have an \textbf{active learning loop} \cite{activesql} that can make training more efficient.

\begin{figure}[h!]
\centerline{\includegraphics[width=0.7\linewidth]{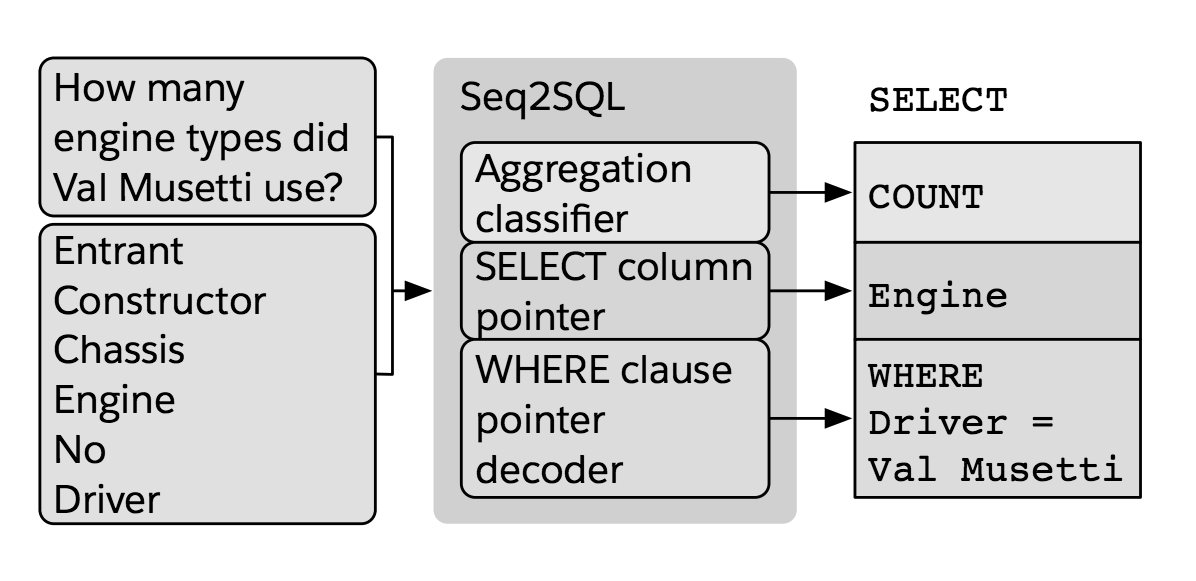}}
\caption{\textit{Seq2SQL \cite{seqsql} uses a pointer network to select question words and column names in order to construct an SQL query.}}
\end{figure}

Program induction/synthesis tasks that require query parsing are by far the hardest to succeed at. It isn't that language is hard to parse, since there are numerous papers which involve question answering (e.g. \cite{vqa}). Rather, there is simply the additional challenge of gathering enough high quality data for the program synthesis tasks. It is expensive to curate pairs of (natural language, program language) or (program language, program language) on which to train a deep learning model. While crowdsourcing answers is possible, it does not make for a cohesive, comprehensive and rigorous dataset. Until the deep learning and/or program synthesis communities launch a large scale effort to create large-scale datasets like ImageNet \cite{imagenet}, this class of problems will remain very challenging. 

\section{Strategies to Reshape Program Induction}
It may be surprising to know that even for a field as new as neural program induction, fundamental model mechanisms have already been dismantled and reimagined. The following concepts are examples of these reformulations that serve to optimize and complete certain aspects of program induction. 

\subsection{Structured Attention}
While attention has proven to be an important tool for deep learning, the approach has more promise yet. In Structured Attention Networks \cite{struct-atten}, the researchers attempt to dismantle some assumed limitations of attention mechanisms. 

The paper makes the key observation that soft attention mechanisms are designed to attend over \textit{individual components} in a set. It would be of tremendous value to attend over \textit{entire groups or segments} of a set collectively since they may be correlated or anti-correlated. By learning this inherent relationship structure in the set and then attending on the structure, attention becomes more effective. 

The paper describes a mechanism to do this using a mathematical feature called a Linear Conditional Random Field (LCRF). It is similar to a Markov Chain in that it connects entities and assumes dependencies only on a pairwise basis. By forming \textit{cliques in a graph} consisting of all a set's objects, one can attend to those structures collectively. They implement this as a new neural network layer type, and so can theoretically be used as a stand-in replacement for any current attention use. It therefore stands to be significant for practically all neural program induction models, since they rely on attention to govern various processes. 

\subsection{Hierarchical Memory}
Many of the neural program synthesis models have a setup in which a recurrent core acts as a controller and interacts with external memory. In "Hierarchical Attentive Memory" \cite{ham}, the researchers specifically attempt to optimize the efficiency and usability of this memory structure. Consider the memory bank of the NTM, which uses content based addressing. To perform hard selections, attention must be done on each of the memory indices, i.e. $\mathcal{O}(n)$ to find the maximum similarity.

Inspired by canonical data structures, the implementation of the memory unit takes the form of a binary tree, which has access time $\mathcal{O}(\log n)$. Content-based attention is done sequentially as a tree traversal, with a threshold of similarity dictating whether to traverse to the left or right child node. Initializing the structure consists of an embedding operation on all the leaf nodes, and then join operations to merge nodes while keeping the same dimensionality. Memory writes result in update operations which cascade upward throughout the tree hierarchy back to the root node.

The researchers demonstrate that the framework is adaptive enough to completely simulate common data structures such as double-ended queues, stacks and heaps when used without an RNN controller. When a controller utilizes the memory, it succeeds at programming tasks far better than without. The research stands out for providing a practical way to give complex models an easier time in training, much like the well-known strategies of dropout and batch normalization.

\begin{figure}[h!]
\centerline{\includegraphics[width=\linewidth]{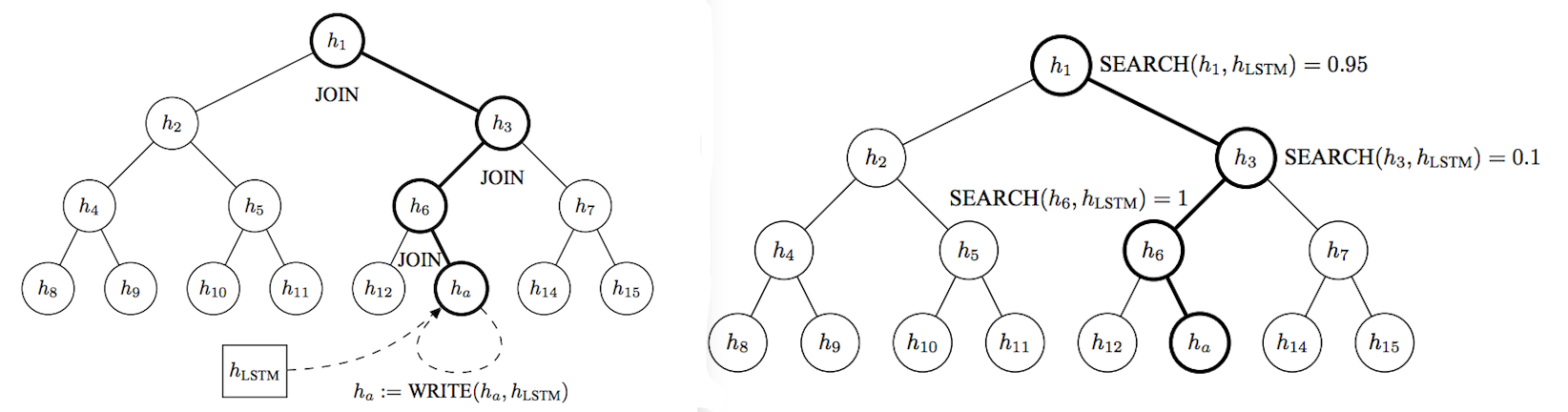}}
\caption{\textit{A write operation updates the parents of the leaf that was written to (Left). A search operation acts just like it would on a binary tree (Right).}}
\end{figure}

\subsection{Enabling Recursion}
\textbf{Recursion }is a fundamental concept taught in algorithms classes. Self-reference is an elegant and powerful tool for expressing regular, hierarchical programs. The strategy also forms the basis of all \textbf{divide-and-conquer} style algorithms, which merge results on smaller versions of the same problem. The neural program synthesis models described above are functions in of themselves and so recursion in this context would mean that the \textit{model calls itself.} For purely connectionist models, this is \textit{impossible}, since the only operations they are capable of are mathematical transformations on their input through tensor arithmetic, nonlinearities and perhaps attention mechanisms. Even then, Neural RAM and Neural Programmer cannot perform true recursion because their modules are predefined, static and \textit{not learned}. In regular computing, recursive programs also necessarily instantiate a new stack frame where local variables may have the same names but take on different values from the caller's frame. At the termination of the program, control is returned to the caller and any overloaded variables have their values restored.  

In "Making Neural Programming Architectures Generalize via Recursion" \cite{recursion}, the researchers acknowledge these issues and state (to their knowledge), the Neural Programmer-Interpreter (NPI) is the only currently existing Neural Programming Architecture (NPA) that could support recursion. The NPI has a catalog of program-key pairs which are learned and can be called with arguments. The straightforward idea of the research is to give the NPI the option to \textbf{call itself} as one of the many functions and supply arguments to it. By introducing the ability to produce recursive calls, an NPA is incentivized to \textbf{learn abstractions} for what the actual task is entailing. The overall length of the program to be emitted by a single instantation of NPA is also significantly \textit{shorter}, and can scale indefinitely because the workload for any given NPA in a stack frame remains constant.  

The experiments demonstrate far better generalization for four algorithmic tasks that are all easily defined recursively. Furthermore, the researchers show that recursive NPAs have \textbf{provably correct generalization}. This validates the researchers' hypothesis that the original implementations of NPI are likely overfitting to insignificant aspects of their problem, such as input length. 

\subsection{Greedy Algorithms}
Like recursion, greedy heuristics are foundational in the study of algorithms. Often, a provably good solution can be obtained by ignoring the notion of edge cases and instead focusing on the immediate options presented in any point of a partial solution. Greedy algorithms find use in \textbf{dynamic programming} solutions, which are characterized by iteratively building up a solution through traversing a directed acyclic graph of \textbf{partial solutions}. For a neural programming model to learn greedy algorithms, there must be a notion of a partial solution as well as a methodology by which to change state to another partial solution. Arguably, encoder-decoder models such as the Pointer Network only function by attempting greedy solutions. However, this explanation of the Pointer Network is dubious at best, since with large problem sizes of the Traveling Salesman Problem, the network is liable to produce \textit{invalid results} with significant probabilities.

In "Learning Combinatorial Optimization Over Graphs" \cite{graphs} the researchers demonstrate high quality results when the neural programming model is constrained to produce greedy algorithms. In particular, \textbf{reinforcement learning} was a potent paradigm for this task since the framework naturally entails notions of "states" and "actions." Here, a state would be a partial solution, and since the action space can be arbitrarily limited at any state, the \textit{available actions can always be made to respect constraints.} Reinforcement learning also entails discounted future rewards, so the learned greedy algorithm can be superior to a traditional one because it is flexible to future partial solutions.

\begin{figure}[h!]
\centerline{\includegraphics[width=\linewidth]{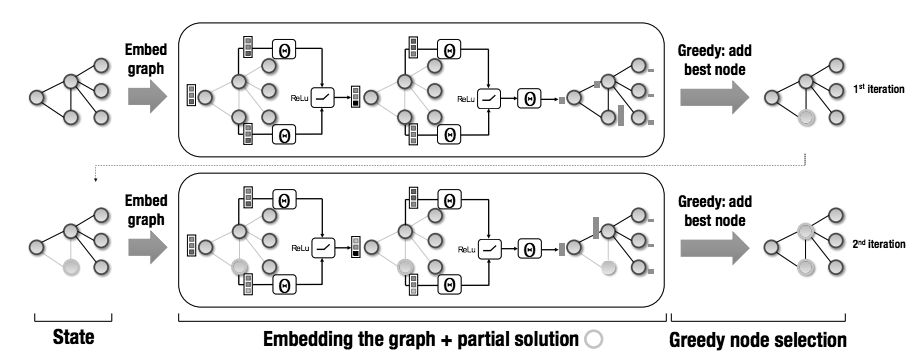}}
\caption{\textit{Reinforcement learning is a suitable mechanism to learn greedy algorithms in a combinatorial optimization context. }}
\end{figure}

The empirical results of this strategy are impressive and elegant. Training on small graph sizes was effective for generalizing to graphs which contains hundreds of times more vertices. This example also shows an example of a \textit{practical example} of neural program induction: producing \textbf{intelligent heuristics} that could potentially outdo man-made ones. 

\section{Neural Program Synthesis}
In the prior sections, we have seen that the field of neural program induction has inspired considerable enthusiasm in the deep learning community. In particular, there are noticeable attempts at building creative architectures (e.g. memory and embedded function hierarchies) that build upon the pitfalls of other styles. Furthermore, there are several publications that try to address and advance the fundamental mechanisms that power these architectures (e.g. recursion, hierarchical memory). This contrasts with with the work done on program synthesis problems, where there is no clear evolution in technique. As we have discussed, program synthesis is a naturally harder problem for deep learning to tackle, so this explains why recent work seems very exploratory and not as iterative. Hence, the following sections are meant only to briefly introduce the different work that has been done so far and some recommendations as to directions the field should invest its efforts into. 

\subsection{Examples of Neural Program Synthesis}
A good place to start the discussion is FlashFill \cite{ffill}, an important non-neural algorithm that works to synthesize programs of regular expression string transformations. The algorithm stands as one of the most often cited use-cases of program synthesis, since it is used in Microsoft Excel to infer programs that define a column as a function of the contents of other columns. String transformations can be summarized with a fairly simple DSL, consisting of concatenations and slicing based on conditions.  

"Neuro-Symbolic Program Synthesis" (NSPS) \cite{nsps} and "Robustfill" \cite{rfill} are two works that use deep learning methods to also tackle the FlashFill problem. NSPS proposes a new architecture called the Recursive Reverse-Recursive Neural Network (R3NN), which can be thought of as an RNN with a tree structure rather than a linear one. The idea is to synthesize a program represented in an \textbf{abstract syntax tree} (AST). The nodes and edges of the R3NN serve this purpose, and the model acts by incrementally growing the tree until it terminates. In "Robustfill," the researchers stick to a sequence to sequence model with attention to construct a program, but make modifications to the DSL so as to increase the vocabulary by making compositional programs into literal ones. In a third work, "Abstract Syntax Networks" \cite{asn}, the researchers also built AST representations of programs but instead for general purpose programming languages. A notable facet of the architecture design is that the decoder which generates the AST is actually composed of several \textit{mutually recursive modules}. Consistent with the different layers of abstraction in imperative programming languages, the active decoder can call upon other modules to create a constructor. This is, in some ways, similar to the sub-function calls used in the Neural Programmer-Interpreter. 

\begin{figure}[h!]
\centerline{\includegraphics[width=\linewidth]{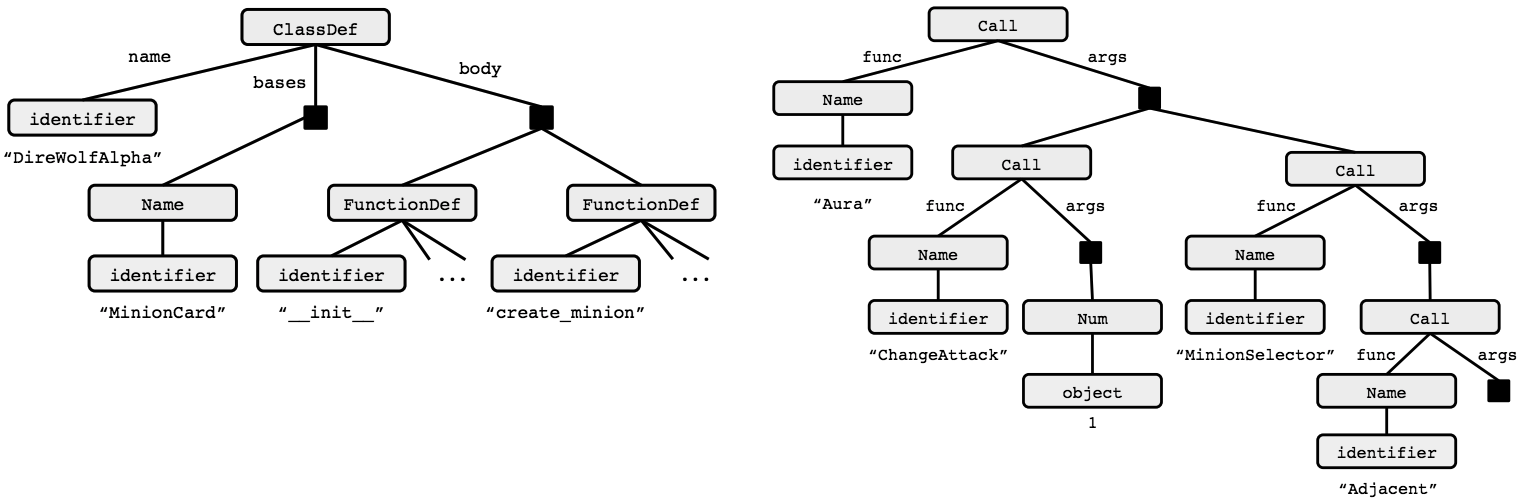}}
\caption{\textit{A program can be suitably represented in an abstract syntax tree \cite{asn}. This, however, requires attention and function hierarchy to be fully effective.}}
\end{figure}

These papers showcase applications of (hierarchical) attention, program embeddings, management of function hierarchy, and constraint satisfaction. These papers were also written in the time since the NPI was published, and so chronologically show a spread of program induction architecture ideas to the program synthesis domain. 

Other fascinating works in neural program synthesis are less connected and show the diversity of the challenges the field is tackling. For example, the "Deepcoder" framework \cite{deepcoder} uses the neural network component to augment and improve the perfomance of traditional program synthesis techniques like SMT solvers. Their objective is to be able to perform well in online coding challenges, and the neural network is used to deduce patterns and insights from the problem specification. There is also the idea of the Differentiable Interpreter \cite{diff-interpreter}, which learns programs by correcting itself when the instructions it chooses are executed at inference time. The differentiable interpreter framework, however, is demonstrated as merely a starting point and a proof of concept, and the literature shows that it has very far to go before being competitive with traditional program synthesis. 

\subsection{Future Research Recommendations}
One may ask what is it even that the field neural program synthesis hopes to achieve as its end goal. An idealist would say that \textit{truly} solving program synthesis is the last programming problem mankind will have to solve. The argument would be that such a solution could synthesize programs of arbitrary complexity better than we can in terms of time spent to achieve program specifications. Then, we simply have that mechanism output another mechanism which is better than itself (since we created the it), and so we get exponential progress and humans never have to program again. 

If that is the vision we wish to align ourselves with, then here are some examples of broad research areas that need further exploration.
\begin{enumerate} 
\item \textbf{Specifically designing neural architectures} to excel at the difficult problems of program synthesis. This is the subject matter that most of this literature review has covered, mostly because it serves as a launchpad to inspire more targeted research in the area. These architectures also need more theory to go along with their results. 
\item We will technically need to solve \textbf{natural language processing} (NLP) too, since a complete program synthesis solution would entail being able to program based off of English instructions or specifications. Processing docstrings of existing codebases would be good for training data in this regard.
\item A greater emphasis should be placed on \textbf{reinforcement learning} techniques. It may be useful to frame the model as an agent which is learning how to program and taught by reward functions. This opens up the possibility of interactions with non-differentiable model components as seen in \cite{xinyun}
\item Deep learning operates on distributed representations of data, and so we will need to continually refine our methods for \textbf{representation learning}. Embeddings of code structures may require special consideration than for semantic embeddings seen in traditional NLP. This is also important for reinforcement learning techniques, since there will be notions of state and action involved in these representations. 
\item To build and modify codebases interpretable by humans, our solution will also ideally synthesis programs with human \textbf{source code bias}. In particular, the structure, concision and conventions of code would ideally be similar to human styles. This also relates to the NLP problem, since naming conventions could fall under this category. An example of favoring human source code bias using traditional program synthesis techniques is in \cite{sampling}.
\item Building more complete solutions will no doubt require greater \textbf{automation in learning.} Current methods rely heavily on hand-crafted curricula to incrementally challenge a model and improve generalization. This may be easier with program induction since the program induced will always be the same, but this is not true in program synthesis. Creating a curriculum of, say, Python code snippets is far harder than increasingly long lists to sort. \cite{autocurr} is an exciting example of automatic curriculum generation.
\item To achieve the best results, neural program synthesis will require more research in \textbf{meta-optimization}. This term refers to the notion of optimizing an optimization procedure. One application is optimizing the learning process over the commonalities of multiple tasks, as seen in \cite{maml}. This form will lead to quick adaptation to any particular task. Another application, seen in \cite{arch-search}, is extending learning to not only change parameter values but also hyperparameters, including the model architecture itself. 
\end{enumerate}

\subsection{Conclusion}
Neural program synthesis is a field with very lofty goals and tremendous latitude for exploration. This literature review sought to first, explain the problem statement, history and intuitive expectations for neural program induction and synthesis. In particular, neural program induction has seen a lot of focus and evolution take place in the methods and problems it seeks to solve. So much so, in fact, that notable work has been done to reshape the foundations of several aspects of program induction techniques. This led us to an overview of work done on program synthesis, which is relatively disorganized and has greater challenges to overcome. With regards to those challenges, we propose some fields of study and work done in them that may hold the key to solving neural program synthesis in the future.

\printbibliography[
heading=bibintoc,
title={References}
]

\end{document}